%% file: ms.tex
\documentclass[11pt]{article}

\usepackage{graphicx} % Required for inserting images
\usepackage[table]{xcolor}
\usepackage{titling}
\usepackage{geometry}
\usepackage{algorithm} 
\usepackage{hyperref}
\usepackage{natbib}
\usepackage{subfiles}
\usepackage{amsmath}
\usepackage{amssymb}

\usepackage{algpseudocode}
\title{DiffFinger: Advancing Synthetic Fingerprint Generation through Denoising Diffusion Probabilistic Models}

\author{
Freddie Grabovski, Lior Yasur, Yaniv Hacmon, Lior Nisimov, Stav Nimrod
}

\date{March 2024}

\begin{document}

\maketitle

\begin{abstract}
This study explores the generation of synthesized fingerprint images using Denoising Diffusion Probabilistic Models (DDPMs). The significant obstacles in collecting real biometric data, such as privacy concerns and the demand for diverse datasets, underscore the imperative for synthetic biometric alternatives that are both realistic and varied. Despite the strides made with Generative Adversarial Networks (GANs) in producing realistic fingerprint images, their limitations prompt us to propose DDPMs as a promising alternative. DDPMs are capable of generating images with increasing clarity and realism while maintaining diversity. Our results reveal that DiffFinger not only competes with authentic training set data in quality but also provides a richer set of biometric data, reflecting true-to-life variability. These findings mark a promising stride in biometric synthesis, showcasing the potential of DDPMs to advance the landscape of fingerprint identification and authentication systems.

\end{abstract}

\section{Introduction}
% Motivation
Fingerprint identification is one of the most common personal identification methods, and it's use is increasing with technology advancement. However, there are significant challenges associated with the collection of real biometric data. Issues such as privacy concerns and the need for datasets that reflect a wide variety of biometric features underscore the importance of developing synthetic alternatives. In response to these challenges, there has been a concerted effort within the scientific community to explore the potential for synthetic biometrics, including fingerprints, aiming to produce results that are not only realistic but also diverse and free from dataset biases \cite{10056846, Maltoni2022}.

% Research Question and Hypothesise
Our research centers on whether Denoising Diffusion Probabilistic Models (DDPMs) can generate high-quality and diverse fingerprint images. Accordingly, the foundational hypothesis of our study is that DDPMs not only have the potential to create fingerprints but also to achieve this with notable quality and variability. This underlying assumption directs our research, emphasizing the potential use of DDPMs to overcome the prevailing challenges in the synthesis of fingerprints, which play a vital role in creating models that can augment the security and dependability of fingerprint identification systems.

% Related Work Overview
The application of DDPMs to the generation of synthetic fingerprints, a project we refer to as DiffFinger, is inspired by the model's proven effectiveness in other biometric fields, such as the imaging of X-rays and MRI scans \cite{Mahaulpatha2024ddpm}. DDPMs introduce an innovative method for image generation that relies on a controlled diffusion process. During the model's training phase, real data is incrementally corrupted with noise following a specific pattern. This is then followed by a denoising phase, where a model, akin to a U-Net, is trained to counteract the noise addition. This process of "backward diffusion" is applied iteratively, refining the initial noise into images that statistically resemble those found within the training dataset \cite{10081412}. For a more in-depth discussion on the mechanics of DDPMs and their application in the synthesis of fingerprints by DiffFinger, readers are directed to the detailed explanation provided in the designated section on diffusion models (Section \ref{sec:diffusion_models}).

To systematically explore this hypothesis, we first aimed to identify and evaluate the range of computational frameworks that could potentially support the generation of synthetic images. Secondly, we delve into identifying state-of-the-art (SOTA) architectures specifically tailored for the synthesis of fingerprint images, this is crucial for understanding the current technological frontier in fingerprints synthesis and for situating DDPMs within this landscape. We find that the field of image synthesis has witnessed notable advancements, particularly with the development and application of Generative Adversarial Networks (GANs). GANs have been celebrated for their ability to generate highly realistic images and have found applications across a wide range of computer vision tasks \cite{IGLESIAS2023100553}, and are currently the state-of-the-art method for synthetic fingerprint generation. Despite their successes, GAN-based methods are not without limitations, including issues related to mode collapse and the instability during the training phase. It is within this context that we propose DDPMs based synthetic fingerprint generator model as a promising alternative.

To ensure that our exploration of DDPMs' capability in generating synthetic fingerprints is both rigorous and meaningful, we survey the metrics used as criteria for success in the synthesis of fingerprint images. These metrics, allow us to establish a clear and objective framework for assessing the effectiveness of DDPMs in producing fingerprint images that are not only realistic and high-quality but also diverse and representative of the variability inherent in real-world biometric data.

In exploring the potential of Denoising Diffuse Probabilistic Models (DDPM) within the realm of biometric identification, we seek to investigate several key questions. Firstly, we examine whether DDPM can be utilized to generate high-quality fingerprints. Secondly, the importance of various pre-processing steps for fingerprints is evaluated, assessing how these processes affect the quality of the generated fingerprints. Finally, we explore the capacity of DDPM to create realistic impressions of fingerprints.

% Contribution and Main Results
Our investigation with Denoising Diffusion Probabilistic Models (DDPMs) via the DiffFinger model has yielded promising outcomes in the domain of synthetic fingerprint generation. Through comprehensive evaluations against existing methods that benefited from training on the NIST SD4 dataset, which is of higher quality than any LivDet variants used for DiffFinger, we have demonstrated the performance of DiffFinger in creating synthetic fingerprints that are both diverse and of high quality (Section \ref{sec:results}). Key findings from our work include the successful synthesis of fingerprints that exceed the quality benchmarks set by the original training datasets, as evidenced by NFIQ2 scores, and the achievement of a closer resemblance to the authentic dataset distributions, as indicated by FID scores (Section \ref{sec:results}). Furthermore, our analysis of dataset diversity through Bozorth3 scores illustrates DiffFinger's capacity to generate fingerprints that offer a broad spectrum of biometric variability. The main contributions of our research are summarized as follows:
\begin{itemize}
    \item Enabled the synthesis of an infinite number of varied and high-quality fingerprint images, demonstrating the potential of DDPMs in generating synthetic fingerprints that can significantly enhance the development and testing of fingerprint identification systems.
    \item Provided the capability to generate multiple synthesized fingerprint impressions for the same identity, facilitating the creation of more comprehensive and realistic fingerprint datasets that can improve the robustness of identification systems against a wide range of samples.
    \item Created a synthesized fingerprints dataset using DiffFinger, which stands as a significant advancement in the field, offering a new resource for researchers and practitioners in biometric security to explore and validate their algorithms.
\end{itemize}

\section{Background}

This section provides an overview of the fundamental terms in the field of fingerprints and the metrics utilized for evaluating the performance of our synthetic fingerprint generation method, DiffFinger. Understanding the intricacies of fingerprint minutiae and the evaluation metrics is essential for understanding the goals and achievements of our approach.

\subsection{Fingerprints Minutiae}
\label{minutiae}
In biometric security, minutiae points serve as features extracted from fingerprints to uniquely identify individuals. Fingerprint minutiae primarily consist of various types of ridge endings and bifurcations. A ridge ending refers to the point where a ridge terminates, and a bifurcation is where a single ridge splits into two separate ridges. Other less common minutiae types include dots or islands (small, isolated segments of ridges), enclosures (a single ridge that bifurcates and then rejoins), and short ridges (ridges that begin and end within a short distance without bifurcating). These features are considered the fingerprints' 'details,' playing a pivotal role in the analysis conducted for forensic purposes, access control systems, and other applications that require identity verification. The process of minutiae extraction and analysis involves several steps. Initially, a fingerprint image is captured and enhanced to increase the clarity of the ridge patterns. Following enhancement, the minutiae points are detected and extracted from the image. This extraction involves identifying the locations and types of minutiae present within the fingerprint. Once minutiae are extracted, they can be represented in a digital format that encapsulates their spatial relationships and orientations, forming a minutiae map that acts as a unique fingerprint signature. The comparison of fingerprint minutiae between two samples forms the basis of fingerprint matching. This process calculates the similarity between the minutiae patterns of the samples, considering factors such as the number of minutiae points that match in type, position, and orientation. A high degree of correspondence between these points indicates a likely match, affirming the identity of the individual \cite{Maltoni2022}.
In our work, we would want to retain the existence of clear minutiae in synthesised fingerprints in order to simulate artificial identity.

\subsection{Metrics}
\label{Metrics}
To assess the quality and diversity of generated fingerprints, we employ three metrics:
\begin{itemize}
    \item \textbf{Fréchet Inception Distance (FID)}: Measures high-level similarity to real fingerprints by comparing their embeddings in a pre-trained image recognition model. Lower FID indicates better quality, implying that generated fingerprints mimic real ones in overall appearance. While the FID  metric has been criticized as a problematic quality measure in recent years \cite{betzalel2022study,chong2020effectively}, we include it for the sake of completeness as it is still widely used in the field.
    
    \item \textbf{Normalized Fingerprint Image Quality (NFIQ2)}: Assesses fine details like clarity, ridge continuity, and minutiae presence. NFIQ2 scores range from 0 to 100, with 100 representing the highest quality and 0 the lowest. A high score thus suggests generated fingerprints resemble real ones in terms of sharpness and realistic minutiae distribution.

    \item \textbf{Pairwise Comparison Scores (Bozorth3 Algorithm)}: Utilizing the Bozorth3 algorithm, we compute similarity scores between fingerprint pairs based on minutiae patterns. Higher scores indicate closer resemblance between fingerprints, reflecting the uniqueness and variability of each fingerprint in our dataset. A Bozorth3 score higher than 40 is usually used to indicate a true identity match \cite{ko2007users}.
\end{itemize}

\section{Related Work}
This section outlines the progression of methods used for generating synthetic fingerprints, transitioning from initial algorithmic approaches to current deep learning models. It briefly describes the evolution of techniques, highlighting key developments and the shift toward using advanced models like Generative Adversarial Networks (GANs) for enhanced realism and privacy. Additionally, advancements in Denoising Diffusion Probabilistic Models (DDPM) across various domains are also detailed.

\subsection{Fingerprint Synthesis}
Fingerprint synthesis has progressed from early algorithmic approaches\cite{UZ2009979} to textural pattern synthesis. Initially, algorithmic methods aimed to manually replicate the intricate patterns and features of fingerprints, focusing on the mathematical modeling of ridges and minutiae. Despite their innovation, these early methods struggled to fully capture the real-world complexity and diversity of fingerprints. Building on this foundation, researchers moved towards textural pattern synthesis techniques, seeking to enhance realism by creating more varied and authentic ridge patterns. Although these advancements led to more realistic-looking fingerprints, the reliance on predefined textural rules still posed limitations on the achievable diversity and authenticity.

\subsubsection{Autoencoders}
The adoption of Autoencoders, particularly Variational Autoencoders (VAEs), represents a notable advancement in fingerprint synthesis. These neural networks compress input data into a condensed representation, then reconstruct it, facilitating the generation of synthetic fingerprint images by capturing the complex distributions of real fingerprint data. This methodology enhances synthesis and reconstruction fidelity by optimizing with Mean Squared Error (MSE) and Structural Similarity Index (SSIM).

Significant contributions to this area include the work by \cite{8914499}, which employs VAEs to create realistic synthetic fingerprints, optimizing for accuracy and fidelity. Another noteworthy effort is \cite{shoshan2023fpgancontrol}, focusing on the generation of fingerprints for improved recognition accuracy in synthetic data training, though it's limited by its dataset scope and exploration depth in computational complexity and generalizability.

\subsubsection{Generative Adversarial Networks}
Generative Adversarial Networks (GANs) have revolutionized fingerprint synthesis by employing a dual-network architecture: the Generator (G) creates data samples from noise, while the Discriminator (D) assesses them against real data. This adversarial dynamic fosters the generation of highly realistic fingerprints by pushing G to produce indistinguishable synthetic data and enhancing D's discernment capabilities. Pioneering contributions in this domain include Goodfellow et al.'s foundational work \cite{goodfellow2014generative}, which laid the groundwork for using GANs in diverse applications, including fingerprint synthesis.

Significant advancements are evidenced in works like Finger-GAN \cite{minaee2018finger} by Shervin Minaee and AmirAli Abdolrashidi, which enhances fingerprint realism through DC-GAN architecture and a total variation regularization, and Bouzaglo et al.'s \cite{bouzaglo2023synthesis} framework that synthesizes and reconstructs fingerprints with an emphasis on privacy, demonstrated by the SynFing dataset, which we will compare our performance to. These efforts validate GANs' ability to produce synthetic fingerprints with high fidelity and diverse characteristics, positioning GANs as a pivotal technology in biometric synthesis and privacy preservation.

\subsubsection{Privacy and Personalization}
Recent advancements in Generative Adversarial Networks (GANs) for fingerprint synthesis have expanded beyond realism to include privacy-focused reconstruction and the generation of multiple images per identity. This progression from algorithmic methods to advanced GANs showcases the evolving landscape of fingerprint synthesis, enhancing not just the authenticity and diversity of synthetic fingerprints but also their application scope, particularly in secure biometric systems. This evolution underscores ongoing innovations aimed at improving synthetic fingerprint quality while addressing ethical and technical challenges in biometrics.

\subsection{Synthesis Using DDPM}

The integration of advanced probabilistic models into medical image synthesis, as highlighted by the MT-DDPM framework in the study by Pan et al. \cite{10185559}, marks a significant advancement in the field, offering a novel approach by incorporating forward diffusion with Gaussian noise into medical images for enhanced denoising efficiency. This method not only demonstrates superior performance over traditional generative models through rigorous evaluations but also underscores the evolving complexity and resource demands associated with cutting-edge generative techniques.

Parallel efforts in synthetic ECG signal generation \cite{10185559} further explore the capabilities and limitations of Denoising Diffusion Probabilistic Models (DDPMs), revealing the potential for improvement in performance metrics compared to the WGAN-GP model, a variant of the Generative Adversarial Network known for its stability and sample diversity. These explorations collectively underscore the dynamic and ongoing refinement in the domain of synthetic signal generation methodologies, pointing to the need for continuous innovation.

Building on this momentum, Wang et al.\cite{wang2022semantic} paper introduces the Semantic Diffusion Model (SDM), utilizing DDPMs for creating high-quality and diverse semantic images, thus marking a substantial leap over traditional GAN methodologies. By innovatively processing semantic layouts and noisy images within a distinct U-Net framework, complemented by multi-layer spatially-adaptive normalization and a classifier-free guidance sampling strategy, the SDM framework significantly amplifies the quality and diversity of image synthesis. This not only achieves state-of-the-art performance across several benchmark datasets but also highlights the broader applicability and potential of DDPMs beyond medical imaging into semantic image synthesis, linking the continuous advancements in probabilistic models with practical applications in diverse imaging domains.

\section{Methodology}

\subsection{Diffusion Probabilistic Models}

\label{sec:diffusion_models}
Denoising Diffusion Probabilistic Models (DDPMs) harness the principles of nonequilibrium thermodynamics to offer a novel generative modeling technique, which we adapt for synthesizing fingerprints in an unprecedented application.

The core of DDPM functionality is divided into forward and backward diffusion phases. The forward phase incrementally applies Gaussian noise to an initial image $x_0$, resulting in a sequence that culminates in a purely noised state $x_T$. This is governed by the formula:
\begin{equation}
    x_{t} = \sqrt{\alpha_{t}} x_0 + (1 - \alpha_{t})^{1/2} \epsilon, \quad \epsilon \sim \mathcal{N}(0, I),
\end{equation}
where $\alpha_{t}$ is a factor that determines the noise level at each step $t$, and $\epsilon$ is the sampled noise.

For the backward or reverse phase, the objective is to reconstruct the original image starting from $x_T$. This process leverages a neural network, modeled as $\epsilon_{\theta}(x_t, t)$, to predict the noise component $\epsilon$ at each timestep and progressively denoise the image. The network optimizes the following objective:
\begin{equation}
    \mathcal{L} = \mathbb{E}_{t,\epsilon}\left[|| \epsilon - \epsilon_{\theta}(x_t, t) ||^2\right],
\end{equation}
where $\mathbb{E}_{t,\epsilon}$ denotes the expectation over all timesteps and noise samples. Through this optimization, the model effectively learns to invert the noise addition process, gradually reconstructing the clean image from its fully noised counterpart.

This advanced approach enables the precise generation of synthetic fingerprints by iteratively refining noise-added images, demonstrating the potential of DDPMs to enrich biometric data generation with high fidelity and diversity.

\subsection{Training details}
Our framework revolves around the Diffusion model, for which we used the default hyper-parameters as outlined in an existing code~\footnote{https://github.com/lucidrains/denoising-diffusion-pytorch} with the exception of the model's dimensions and batch size, which we changed to 32 \texttt{init\_features} and 16 respectively.
Our model was trained on the LivDet dataset, utilizing an RTX 4090 Nvidia GPU. This dataset, amassed by researchers at Clarkson University over time, was obtained through direct request \cite{marcialis2009first, yambay2012livdet, ghiani2013livdet}.

\subsection{Data Preprocessing}
In the preprocessing stage, our objective was to ensure the input fingerprints were of high quality and uniformly formatted to optimize the DDPM's learning process. The preprocessing pipeline consisted of three primary steps: filtering based on fingerprint quality, cropping and centering, and adjusting to a 1:1 aspect ratio.

Firstly, we utilized the NFIQ metric, as described in \ref{Metrics} to filter our dataset. To maintain a high standard, we selected fingerprints with NFIQ scores of 1 or 2, discarding those with inferior quality. This step ensured that our model was trained on clear and detailed fingerprints, which is crucial for generating high-fidelity synthetic prints.

Secondly, we addressed the issue of whitespace in fingerprint scans. Some scans contain significant areas of non-fingerprint space, which could detract from the model's efficiency and output quality. To mitigate this, we implemented a cropping and centering procedure. This process involved detecting the fingerprint's boundaries and adjusting the image so that the fingerprint is centered. The decision to crop was based on a predetermined average pixel value threshold; images exceeding this threshold were cropped to eliminate excess whitespace, focusing the model's attention on the fingerprint itself.

Finally, we adjusted all images to a 1:1 aspect ratio. Aspect ratio standardization was crucial for the DDPM model to take a constant input size while maintaining the structural integrity of the fingerprints, preventing distortion that could lead to unrealistic synthetic prints. By ensuring all input images conformed to this ratio, we facilitated the model's ability to generate prints that are not only high in quality but also realistic in appearance.

Through this preprocessing pipeline, we established a solid foundation for our DDPM to learn from a dataset characterized by high-quality, well-formatted fingerprints. This approach not only enhances the model's performance but also ensures that the synthetic fingerprints it generates are of practical use in various applications, ranging from biometric security to forensic analysis.
 
 \label{data}
In this work, we inspected the impact of the different processing steps in a leave-one-out manner and compared the performance. We will train a DDPM model for each of the datasets and compare them in \ref{Exp 1} 

\begin{figure}
    \centering
    \includegraphics[width=1\linewidth]{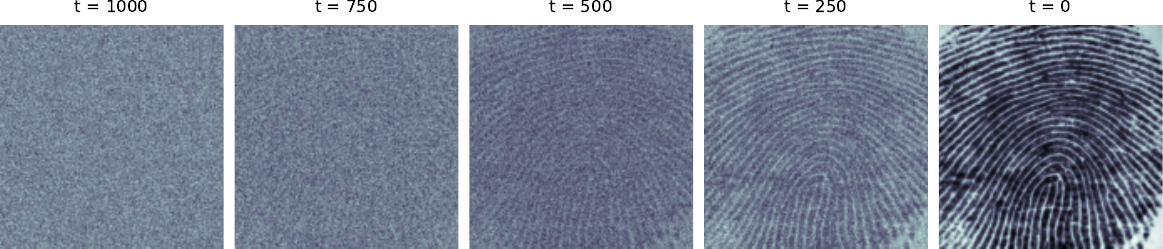}
    \caption{Example of the fingerprint generation process using the backward process of the DDPM}
    \label{fig:DDPM_example}
\end{figure}

\subsection{Inference}

To synthesize high-quality artificial fingerprints, we utilize the trained DDPM-based DiffFinger model, starting from noise $\epsilon\sim \mathcal{N}(0, I)$ as input for the denoising process. This approach involves a backward diffusion process, where we begin with noise and then iteratively estimate $x_{t-1}$ from $x_t$, continuing this process until $t=0$. As a result, we obtain a sample $x_0$ that closely resembles the samples in the training set. This method allows us to generate new images once the neural network's training has converged, demonstrating the capability of our approach in Figure~\ref{fig:DDPM_example}.

\begin{figure}[ht]
    \centering
    \includegraphics[width=0.85\linewidth]{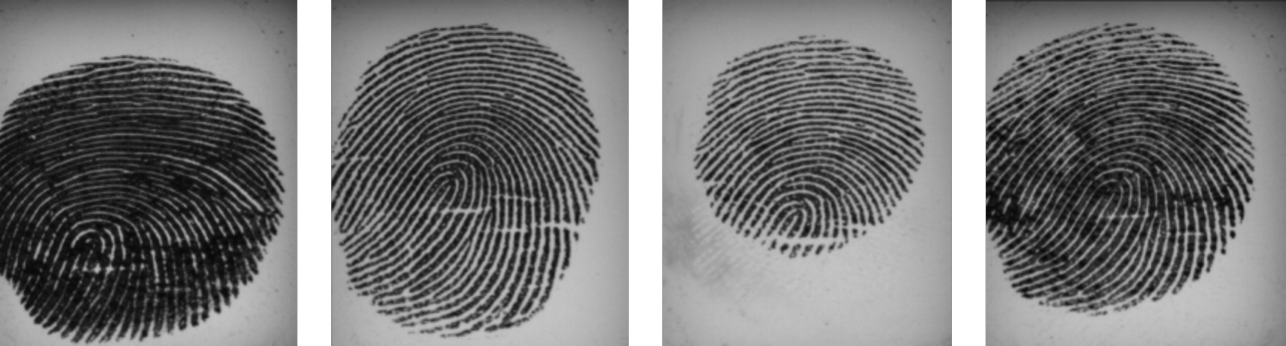}
    \caption{Impressions of the same identity from our training dataset}
    \label{fig:real_impression}
\end{figure}

The proposed DiffFinger method also allows the generation of different impressions of the same identity, the purpose of which is to emulate the real-world which allows fingerprint recognition better performance and robustness. An example of a set impression of the same identity from our training set can be found in Figure~\ref{fig:real_impression}.
To create these impressions, we exploited the capabilities of our DDPM-based model in a unique manner. This procedure aimed to generate sets of fingerprints that, while distinct, represent the same identity. The procedure involves utilizing the backward diffusion process of DDPM in a novel way to achieve this goal.

The procedure begins with initializing a purely noise-based image (at time step  $t=1000$) as our starting point. We then partially denoise this image up to a specific, predefined time step $d$, stopping before reaching the final reconstruction step. This intermediate image, still partially obscured by noise, establishes a unique fingerprint identity. To generate multiple impressions of this identity, we then continued the denoising process from $t=d$ to $t=0$ multiple times. Due to the stochastic nature of the DDPM, each iteration results in a slightly different image, thereby creating variations of the same fingerprint identity.

\subsection{Qualities and Contributions}

This section highlights the advantages of our DDPM-based approach for fingerprint generation compared to the limitations of Generative Adversarial Networks (GANs).

Overcoming Mode Collapse: A critical requirement for our methodology is ensuring diverse and realistic fingerprint generation. However, GANs are susceptible to mode collapse \cite{zhang2018convergence}, leading them to predominantly generate samples from specific regions of the training data distribution. In contrast, DDPMs inherently avoid mode collapse due to their noise-driven training process \cite{bayat2023a}. This enables us to generate a broader range of fingerprints encompassing various acquisition methods (rolled, pressed) and scan types (digital, ink-based), fulfilling a crucial requirement for our methodology.

Enhancing Realism and Capturing Variability: Another key requirement is generating fingerprints with high realism and capturing the inherent variability of real-world data. DDPMs excel in this aspect by iteratively refining noise guided by the learned data distribution. This results in statistically similar samples with improved realism, particularly noticeable in intricate details like ridges and minutiae. Compared to GANs, DDPMs demonstrate superior ability to model these fine-grained features, leading to more realistic and diverse fingerprint generation \cite{bayat2023a}.

Potential for Explainability:  Explainability is increasingly valued in machine learning models. The diffusion process inherent to DDPMs offers unique potential for understanding the model's decision-making. By analyzing the noise prediction and removal steps, we can gain insights into the factors influencing the generated fingerprints. This potential for explainability aligns with the growing emphasis on interpretable models in various applications.

\section{Experiments}
In the following sections we will discuss the experiments conducted to answer our research questions:
\begin{enumerate}
\item Can DDPM be used to create high quality fingerprints?
\item How important are different pre-processing steps for fingerprints?
\item Can DDPM be leveraged to create realistic impressions of fingerprints?
\end{enumerate}

For these experiments, we utilized an Nvidia 4090 RTX graphics card. We also incorporated Nist's Bozorth3 code\footnote{https://github.com/lessandro/nbis/tree/master/bozorth3} and NFIQ2\footnote{https://www.nist.gov/services-resources/software/nfiq-2}, alongside the PyTorch library.

\subsection{Metrics}
In our evaluation of the generated fingerprints' quality and diversity, we refer to three metrics previously detailed in Section~\ref{Metrics}: Fréchet Inception Distance (FID), Normalized Fingerprint Image Quality (NFIQ2), and Pairwise Comparison Scores (Bozorth3 Algorithm).

\subsection{Experiment 1: Populate Dataset}
\label{Exp 1}

In Experiment 1, our objective was to synthesize new, high-quality fingerprints to populate datasets through the application of Denoising Diffusion Probabilistic Models (DDPMs). To systematically evaluate the impact of different preprocessing techniques on the quality of the generated fingerprints, we constructed three distinct datasets, each with a different variant of our preprocessing protocol, derived from the LivDet dataset.

\begin{enumerate}
    \item The first dataset was processed using the entire preprocessing pipeline, incorporating all stages. This variation will be denoted as FP for full pipeline
    \item The second dataset omitted the cropping phase, allowing us to assess the influence of this step on the overall quality of the generated fingerprints.
    \item The third dataset bypassed the initial pre-filtering stage, providing insight into the effect of filtering on the fidelity of the synthetic fingerprints.
\end{enumerate}

Each DiffFinger model was trained exclusively on one of these datasets, and 40,000 synthetic images were created by each model, facilitating a nuanced ablation study of our preprocessing methodologies. Furthermore, to contextualize the performance and quality of our synthetic fingerprints, we conducted a comparative analysis against three established fingerprint synthesis methods:

\begin{enumerate}
    \item The technique devised by Cao and Jain, which leverages minutiae patterns for the generation of synthetic fingerprints, offering a structured approach to fingerprint synthesis \cite{cao2018fingerprint}.
    \item The strategy proposed by Minaee et al., which harnesses the capabilities of deep learning architectures to create realistic fingerprint images \cite{minaee2018finger}.
    \item The SynFing methodology introduced by Bouzaglo et al., representing a contemporary breakthrough in the domain of synthetic fingerprint generation \cite{bouzaglo2023synthesis}.
\end{enumerate}

This experimental setup not only allowed us to dissect the contributions of individual preprocessing steps to the overall quality of the generated fingerprints but also to position our DDPM-based approach within the broader landscape of synthetic fingerprint generation technologies.

Our hypothesis is that the models will achieve both high quality (low FID, high NFIQ2) and high diversity (many low Bozorth3 scores) through our DDPM-based method, demonstrating its ability to generate realistic, detailed, and diverse fingerprints applicable to various scenarios, and that the FP dataset will result in the highest quality images.

\subsection{Experiment 2: Generating Impressions}

In this experiment, we assessed DiffFinger's capability to produce fingerprint impressions by employing BOZORTH3 similarity scores. Utilizing the FP model, which performed the best in previous test, we generated 16,000 impressions across 1,000 identities and analyzed the similarity score distribution for pairs of synthetic fingerprints of the same identity, against that of real fingerprint pairs from the training set. Significantly lower scores for synthetic pairs would suggest insufficient identity similarity, while markedly higher scores could indicate a lack of diversity among the impressions.

\section{Results}
\label{sec:results}

This section presents the outcomes of our evaluation of DiffFinger's performance in synthetic fingerprint generation. We begin with a comparative analysis of the image quality produced by DiffFinger against established methods in the field. Following this, we assess the ability of DiffFinger to generate diverse and realistic fingerprint impressions. Each subsection provides detailed insights into these aspects, underlining the advancements our approach offers to the domain of synthetic biometrics.

\subsection{Baseline Comparison}

This subsection focuses on evaluating the performance of DiffFinger in creating synthetic fingerprints. We use two key measures for this evaluation: NFIQ2 scores and the Fréchet Inception Distance (FID). The NFIQ2 scores help assess the quality of the synthetic fingerprints produced by DiffFinger and FID scores measure how similar the collection of synthetic fingerprints is to the original dataset in terms of distribution.

To present our findings, we have prepared two tables: one for FID scores and another for NFIQ2 scores. The FID table compares different methods and their respective training datasets, showing how each method's approach to generating synthetic fingerprints reflects on the similarity of their distribution to the original dataset. The NFIQ2 table, meanwhile, lists the NFIQ2 scores for each dataset, whether synthetic (CaoJain, SynFing and Diffinger) or real (NIST SD4 and LivDet). The detailed results of our evaluation can be found in Table~\ref{tab:fid_scores} for FID scores and Table~\ref{tab:nfiq2_scores} for NFIQ2 scores.
To present our findings, we have prepared two tables: one for FID scores and another for NFIQ2 scores. The FID table compares different methods and their respective training datasets, showing how each method's approach to generating synthetic fingerprints reflects on the similarity of their distribution to the original dataset. The NFIQ2 table, meanwhile, lists the NFIQ2 scores for each dataset, whether synthetic (CaoJain, SynFing or ) or real. The detailed results of our evaluation can be found in Table~\ref{tab:fid_scores} for FID scores and Table~\ref{tab:nfiq2_scores} for NFIQ2 scores.

\subsubsection{Comparative Analysis of NFIQ2 Scores}

 into the NFIQ2 scores (refer to Table~\ref{tab:nfiq2_scores}) .A deep inspectionunderscores a notable achievement of DiffFinger's - creating synthetic datasets that in most cases surpass the quality of images from it's training dataset, which were marked by lower NFIQ2 scores. This observation is critical, especially when considering that competitors such as CaoJain and SynFing, despite achieving high NFIQ2 scores, benefited from training on the NIST SD4 dataset—a dataset of higher quality than any LivDet variants used for DiffFinger.

The comparative advantage of DiffFinger becomes apparent in this context. Despite the lower initial quality of its training datasets, DiffFinger generated fingerprints that exhibit higher NFIQ2 scores. This suggests a potential for DiffFinger to achieve even greater quality improvements if it had access to training data of comparable quality to NIST SD4, positioning it to potentially surpass SynFing and compete closely with CaoJain in terms of fingerprint quality.

Comparing the synthetic datasets created by DiffFinger, training the model on the FP dataset resulted in significantly higher quality images compared to both the No Crop and the No Filter datasets, with a p-value < 0.0001 for both.

Another important insight is the impact of the different preprocessing steps. While the quality of the dataset that underwent a complete preprocessing pipeline had the overall best quality and yielded the better synthetic dataset, an important observation to make is that omitting filtering stage dropped the quality of the fingerprint images considerably more than omitting the cropping phase, emphasising the importance of curating a dataset of clear and identifiable fingerprints for the purpose of training a generative model.

\subsubsection{Implications of FID Scores}

The evaluation of FID scores (see Table~\ref{tab:fid_scores}) further enriches our understanding. FID scores, which measure the similarity between the distribution of synthetic and original datasets, reveal that divergence from the training data's distribution is not inherently detrimental, especially given the lower quality of DiffFinger's training data. This is a critical insight, as it suggests that a higher FID score, indicating greater difference, might not necessarily reflect negatively on the utility of the synthetic fingerprints generated by DiffFinger.

Moreover, for the full pipeline variant of DiffFinger's dataset (FP), which showcased improved NFIQ2 scores, a closer FID score to the authentic dataset was observed, surpassing CaoJain. This indicates a pivotal trend: as the quality of the original training data improves, the synthetic fingerprints generated by DiffFinger are likely to exhibit a closer resemblance to the original dataset in terms of distribution. This observation not only highlights the adaptability of DiffFinger but also its capability to refine the synthetic output in alignment with the quality of the training data.

In conclusion, the comprehensive analysis of both NFIQ2 and FID scores underscores the innovative potential of DiffFinger in the realm of synthetic fingerprint generation. By effectively enhancing fingerprint quality beyond the limitations of its training datasets and indicating a capacity to closely mirror authentic data distributions with higher-quality training data.
\input{Tables/Nfiq_fid}

\subsubsection{Comparative Analysis of Diffinger's dataset Diversity}
Our evaluation of Bozorth3 scores allows comparison between the diversity of the original datasets, and the diversity of the synthetic datasets created by DiffFinger. 
A lower Bozorth3 scores indicates a higher diversity of identities in the dataset.

Since our goal is to create a dataset that will be useful to future research, we have omitted scores of 0 from both the original and synthetic datasets, as such scores are usually caused by bad samples, which can be easily filtered and would be excluded from a real published dataset to preserve it's quality. The filtered figure thus shows the actual diversity of each dataset. A plot of the unfiltered datasets is available in the supplementary material.

Figure~\ref{fig:bozorth} shows that for the FP and No Filter datasets, the mean Bozorth3 scores of images created by DiffFinger are lower than the scores of the original images with p-value < 0.0001, meaning the fingerprints created by our method are more diverse. For the original No Crop dataset, the low mean Bozorth3 score is likely caused by the low quality of the original images, which makes them very different from each other, and it is lower than the mean for the DiffFinger No Crop with with p-value < 0.0001. 

\begin{figure}[ht]
    \centering
    \includegraphics[scale = 0.6]
    {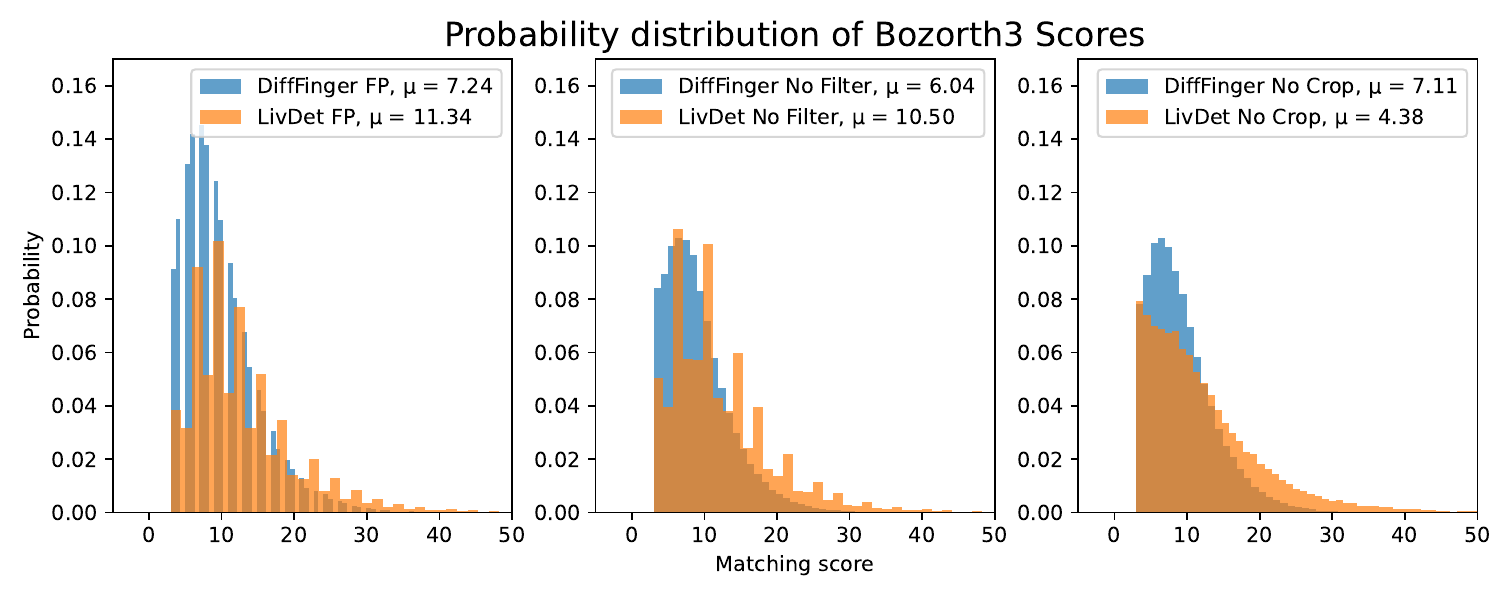}
    \caption{A comparison of the Bozorth3 scores distribution for each training set to it's synthetic counterpart}
    \label{fig:bozorth}
    \vspace{-1em} % Adjust the negative space to bring the following figure closer
\end{figure}
\subsection{Fingerprint Impressions}
\begin{figure}[ht]
    \centering
    \includegraphics[scale=0.6]
    {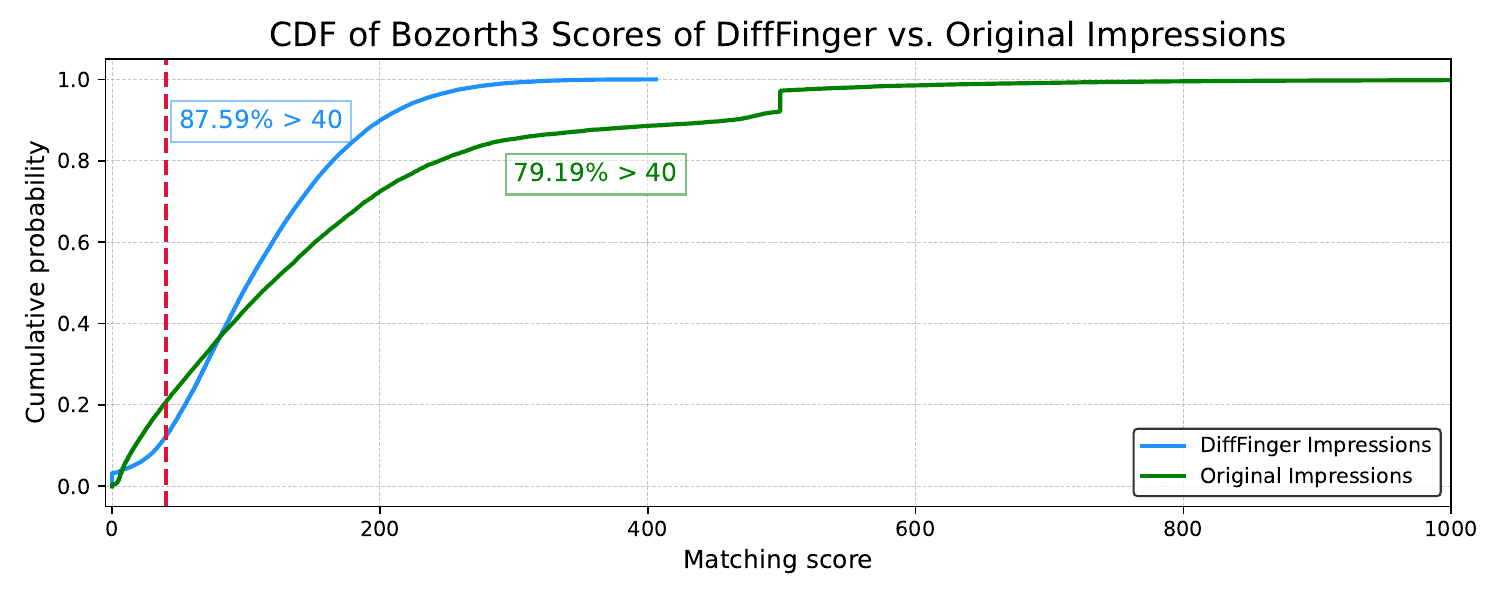}
    \caption{The cumulative density function values of Bozorth3 scores for original vs synthetic impressions, scores over 40 indicate the same identity}
    \label{fig:impression}
    % Optionally, adjust spacing after this figure if needed
\end{figure}
In this section, we evaluate our generated impression's similarity using Bozorth3 scores. Figure~\ref{fig:impression} illustrates the efficacy of the DiffFinger method compared to the original dataset's impressions. The figure shows that 87.59\% of the synthetic impressions surpassed the identity preservation threshold of 40, which means 87.59\% of the impressions successfully maintain the same identity, outperforming the 79.19\% benchmark set by LivDet's authentic impressions. Moreover, while approximately 10\% of genuine impressions scored above 400, indicating near-identical similarity, all synthetic impressions scored below 420. This indicates that DiffFinger can not only produces a higher volume of valid impressions but also ensures a greater diversity of impressions for each identity.

\section{Conclusions and Future Work}
Our study presents DiffFinger, a new approach for synthetic fingerprint generation using Denoising Diffusion Probabilistic Models (DDPMs), offering a novel solution to the challenges inherent in collecting real biometric data. Through the development and analysis of DiffFinger, we have demonstrated the capability of DDPMs to produce high-quality and diverse fingerprints, significantly advancing the field of biometric synthesis. Our findings affirm the hypothesis that DDPMs can effectively synthesize fingerprints that surpass the quality of their training datasets and closely mimic the distribution of authentic fingerprint data, as evidenced by NFIQ2, FID, and Bozorth3 scores.

While our research has taken significant strides in synthetic fingerprint generation, the exploration of DDPMs in this domain is far from complete. Future work can extend in several promising directions:
\begin{itemize}
    \item \textbf{Exploring Advanced Architectures:} Investigating more sophisticated or alternative DDPM architectures could further improve the quality and efficiency of synthetic fingerprint generation. Advancements in neural network design and training methodologies may yield even more realistic and varied fingerprints.
    \item \textbf{Cross-Domain Applications:} Applying the principles of DDPM-based fingerprint synthesis to other biometric modalities, such as facial recognition or iris scans, could open new avenues for comprehensive biometric security solutions. Exploring these applications could significantly impact the broader field of synthetic biometrics.
    \item \textbf{Training on Higher Quality Datasets:} Enhancing the training of DiffFinger with datasets of superior quality could significantly refine the fidelity and diversity of the generated fingerprints. Access to more detailed and diverse training data would likely yield synthetic fingerprints that are even closer to the intricacies and variations found in real biometric data, thereby improving the model's utility and applicability.
    \item \textbf{Ethical and Legal Considerations:} As synthetic biometric technologies evolve, so too should the ethical and legal frameworks that govern their use. Future research should include a focus on the implications of synthetic biometrics in privacy, security, and regulatory compliance.
    \item \textbf{Integration with Real-world Systems:} Pilot studies integrating DDPM-generated fingerprints into operational biometric systems could provide valuable insights into their practical application and performance in real-world scenarios. Such studies would also help identify areas for further improvement in synthesis techniques.
\end{itemize}

\bibliographystyle{unsrtnat}

\bibliography{references}

\end{document}

%% file: Tables/Nfiq_fid.tex
\begin{table}[ht]
\hspace*{-5mm} % Adjust the negative value as needed to shift to the left
\begin{minipage}[t]{0.5\textwidth}
\centering
\begin{tabular}{ccc}
\hline\hline
\textbf{Dataset}  & \textbf{Mean} & \textbf{Std. Dev.} \\ \hline
NIST SD4          & 44.66         & 17.6               \\
CaoJain           & 61.42         & 14.55              \\
SynFing           & 41.70         & 16.10              \\ \hline
LivDet No Crop           & 14.95         & 14.41              \\
LivDet No Filter       & 25.04         & 13.27              \\
LivDet FP     & 27.47         & 12.27              \\
\rowcolor{gray!20}DiffFinger No Crop      & 19.84         & 12.42              \\
\rowcolor{gray!20}DiffFinger No Filter    & 24.48         & 16.14              \\
\rowcolor{gray!20}DiffFinger FP & 29.36         & 17.63              \\
\hline\hline
\end{tabular}
\caption{NFIQ2 Scores}
\label{tab:nfiq2_scores}
\end{minipage}\hfill
\begin{minipage}[t]{0.5\textwidth}
\centering
\begin{tabular}{ccc}
\hline\hline
\textbf{Method} & \textbf{Dataset} & \textbf{FID Score} \\ \hline
Cao and Jain & NIST SD4 & 113.82 \\
Minaee et al. & FVC 2006 & 70.55 \\
SynFing & NIST SD4 & 6.14 \\
\rowcolor{gray!20}DiffFinger No Crop & LivDet No Crop & 162.71 \\
\rowcolor{gray!20}DiffFinger No Filter & LivDet No Filter & 150.80 \\
\rowcolor{gray!20}DiffFinger FP & LivDet FP & 103.10 \\
\hline\hline
\end{tabular}
\caption{FID Scores}
\label{tab:fid_scores}
\end{minipage}
\end{table}

%% file: ms.bbl
\begin{thebibliography}{22}
\providecommand{\natexlab}[1]{#1}
\providecommand{\url}[1]{\texttt{#1}}
\expandafter\ifx\csname urlstyle\endcsname\relax
  \providecommand{\doi}[1]{doi: #1}\else
  \providecommand{\doi}{doi: \begingroup \urlstyle{rm}\Url}\fi

\bibitem[Makrushin et~al.(2023)Makrushin, Uhl, and Dittmann]{10056846}
Andrey Makrushin, Andreas Uhl, and Jana Dittmann.
\newblock A survey on synthetic biometrics: Fingerprint, face, iris and vascular patterns.
\newblock \emph{IEEE Access}, 11:\penalty0 33887--33899, 2023.
\newblock \doi{10.1109/ACCESS.2023.3250852}.

\bibitem[Maltoni et~al.(2022)Maltoni, Maio, Jain, and Feng]{Maltoni2022}
Davide Maltoni, Dario Maio, Anil~K. Jain, and Jianjiang Feng.
\newblock \emph{Fingerprint Synthesis}, pages 385--426.
\newblock Springer International Publishing, Cham, 2022.
\newblock ISBN 978-3-030-83624-5.
\newblock \doi{10.1007/978-3-030-83624-5_7}.
\newblock URL \url{https://doi.org/10.1007/978-3-030-83624-5_7}.

\bibitem[Mahaulpatha and Abeywardane(2024)]{Mahaulpatha2024ddpm}
Praveen Mahaulpatha and Thulana Abeywardane.
\newblock Ddpm based x-ray image synthesizer.
\newblock \url{https://synthical.com/article/54980370-1e7b-4676-a84b-b15cd9e66243}, 0 2024.

\bibitem[Croitoru et~al.(2023)Croitoru, Hondru, Ionescu, and Shah]{10081412}
Florinel-Alin Croitoru, Vlad Hondru, Radu~Tudor Ionescu, and Mubarak Shah.
\newblock Diffusion models in vision: A survey.
\newblock \emph{IEEE Transactions on Pattern Analysis and Machine Intelligence}, 45\penalty0 (9):\penalty0 10850--10869, 2023.
\newblock \doi{10.1109/TPAMI.2023.3261988}.

\bibitem[Iglesias et~al.(2023)Iglesias, Talavera, and Díaz-Álvarez]{IGLESIAS2023100553}
Guillermo Iglesias, Edgar Talavera, and Alberto Díaz-Álvarez.
\newblock A survey on gans for computer vision: Recent research, analysis and taxonomy.
\newblock \emph{Computer Science Review}, 48:\penalty0 100553, 2023.
\newblock ISSN 1574-0137.
\newblock \doi{https://doi.org/10.1016/j.cosrev.2023.100553}.
\newblock URL \url{https://www.sciencedirect.com/science/article/pii/S1574013723000205}.

\bibitem[Betzalel et~al.(2022)Betzalel, Penso, Navon, and Fetaya]{betzalel2022study}
Eyal Betzalel, Coby Penso, Aviv Navon, and Ethan Fetaya.
\newblock A study on the evaluation of generative models, 2022.

\bibitem[Chong and Forsyth(2020)]{chong2020effectively}
Min~Jin Chong and David Forsyth.
\newblock Effectively unbiased fid and inception score and where to find them, 2020.

\bibitem[Ko(2007)]{ko2007users}
Kenneth Ko.
\newblock Users guide to export controlled distribution of nist biometric image software (nbis-ec).
\newblock 2007.

\bibitem[Uz et~al.(2009)Uz, Bebis, Erol, and Prabhakar]{UZ2009979}
Tamer Uz, George Bebis, Ali Erol, and Salil Prabhakar.
\newblock Minutiae-based template synthesis and matching for fingerprint authentication.
\newblock \emph{Computer Vision and Image Understanding}, 113\penalty0 (9):\penalty0 979--992, 2009.
\newblock ISSN 1077-3142.
\newblock \doi{https://doi.org/10.1016/j.cviu.2009.04.002}.
\newblock URL \url{https://www.sciencedirect.com/science/article/pii/S1077314209000794}.

\bibitem[Attia et~al.(2019)Attia, Attia, Iskander, Saleh, Nahavandi, Abobakr, Hossny, and Nahavandi]{8914499}
Mohamed Attia, MennattAllah~H. Attia, Julie Iskander, Khaled Saleh, Darius Nahavandi, Ahmed Abobakr, Mohammed Hossny, and Saeid Nahavandi.
\newblock Fingerprint synthesis via latent space representation.
\newblock In \emph{2019 IEEE International Conference on Systems, Man and Cybernetics (SMC)}, pages 1855--1861, 2019.
\newblock \doi{10.1109/SMC.2019.8914499}.

\bibitem[Shoshan et~al.(2023)Shoshan, Bhonker, Baruch, Nizan, Kviatkovsky, Engelsma, Aggarwal, and Medioni]{shoshan2023fpgancontrol}
Alon Shoshan, Nadav Bhonker, Emanuel~Ben Baruch, Ori Nizan, Igor Kviatkovsky, Joshua Engelsma, Manoj Aggarwal, and Gerard Medioni.
\newblock Fpgan-control: A controllable fingerprint generator for training with synthetic data, 2023.

\bibitem[Goodfellow et~al.(2014)Goodfellow, Pouget-Abadie, Mirza, Xu, Warde-Farley, Ozair, Courville, and Bengio]{goodfellow2014generative}
Ian~J. Goodfellow, Jean Pouget-Abadie, Mehdi Mirza, Bing Xu, David Warde-Farley, Sherjil Ozair, Aaron Courville, and Yoshua Bengio.
\newblock Generative adversarial networks, 2014.

\bibitem[Minaee and Abdolrashidi(2018)]{minaee2018finger}
Shervin Minaee and Amirali Abdolrashidi.
\newblock Finger-gan: Generating realistic fingerprint images using connectivity imposed gan.(2018).
\newblock \emph{arXiv preprint arXiv:1812.10482}, 2018.

\bibitem[Bouzaglo and Keller(2023)]{bouzaglo2023synthesis}
Rafael Bouzaglo and Yosi Keller.
\newblock Synthesis and reconstruction of fingerprints using generative adversarial networks, 2023.

\bibitem[Adib et~al.(2023)Adib, Fernandez, Afghah, and Prevost]{10185559}
Edmonmd Adib, Amanda~S. Fernandez, Fatemeh Afghah, and John~J. Prevost.
\newblock Synthetic ecg signal generation using probabilistic diffusion models.
\newblock \emph{IEEE Access}, 11:\penalty0 75818--75828, 2023.
\newblock \doi{10.1109/ACCESS.2023.3296542}.

\bibitem[Wang et~al.(2022)Wang, Bao, Zhou, Chen, Chen, Yuan, and Li]{wang2022semantic}
Weilun Wang, Jianmin Bao, Wengang Zhou, Dongdong Chen, Dong Chen, Lu~Yuan, and Houqiang Li.
\newblock Semantic image synthesis via diffusion models, 2022.

\bibitem[Marcialis et~al.(2009)Marcialis, Lewicke, Tan, Coli, Grimberg, Congiu, Tidu, Roli, and Schuckers]{marcialis2009first}
Gian~Luca Marcialis, Aaron Lewicke, Bozhao Tan, Pietro Coli, Dominic Grimberg, Alberto Congiu, Alessandra Tidu, Fabio Roli, and Stephanie Schuckers.
\newblock First international fingerprint liveness detection competition—livdet 2009.
\newblock In \emph{Image Analysis and Processing--ICIAP 2009: 15th International Conference Vietri sul Mare, Italy, September 8-11, 2009 Proceedings 15}, pages 12--23. Springer, 2009.

\bibitem[Yambay et~al.(2012)Yambay, Ghiani, Denti, Marcialis, Roli, and Schuckers]{yambay2012livdet}
David Yambay, Luca Ghiani, Paolo Denti, Gian~Luca Marcialis, Fabio Roli, and S~Schuckers.
\newblock Livdet 2011—fingerprint liveness detection competition 2011.
\newblock In \emph{2012 5th IAPR international conference on biometrics (ICB)}, pages 208--215. IEEE, 2012.

\bibitem[Ghiani et~al.(2013)Ghiani, Yambay, Mura, Tocco, Marcialis, Roli, and Schuckcrs]{ghiani2013livdet}
Luca Ghiani, David Yambay, Valerio Mura, Simona Tocco, Gian~Luca Marcialis, Fabio Roli, and Stephanie Schuckcrs.
\newblock Livdet 2013 fingerprint liveness detection competition 2013.
\newblock In \emph{2013 international conference on biometrics (ICB)}, pages 1--6. IEEE, 2013.

\bibitem[Zhang et~al.(2018)Zhang, Li, and Yu]{zhang2018convergence}
Zhaoyu Zhang, Mengyan Li, and Jun Yu.
\newblock On the convergence and mode collapse of gan.
\newblock In \emph{SIGGRAPH Asia 2018 Technical Briefs}, pages 1--4. 2018.

\bibitem[Bayat(2023)]{bayat2023a}
Reza Bayat.
\newblock A study on sample diversity in generative models: {GAN}s vs. diffusion models, 2023.
\newblock URL \url{https://openreview.net/forum?id=BQpCuJoMykZ}.

\bibitem[Cao and Jain(2018)]{cao2018fingerprint}
Kai Cao and Anil Jain.
\newblock Fingerprint synthesis: Evaluating fingerprint search at scale.
\newblock In \emph{2018 International Conference on Biometrics (ICB)}, pages 31--38. IEEE, 2018.

\end{thebibliography}
